\pdfoutput=1

\documentclass[11pt]{article}

\usepackage{ACL2023}
\usepackage{nopageno} 

\usepackage{times}
\usepackage{latexsym}

\usepackage[T1]{fontenc}

\usepackage[utf8]{inputenc}

\usepackage{microtype}

\usepackage{inconsolata}

\usepackage{graphicx}
\usepackage{multirow}
\usepackage{placeins}
\usepackage{amsmath}
\usepackage{balance}
\usepackage{fancyvrb}
\usepackage{lipsum}
\usepackage{tcolorbox}
\usepackage{float}
\usepackage{listings} 
\usepackage{caption} 
\usepackage{booktabs} 
\usepackage{enumitem}

\usepackage{tikz}
\usetikzlibrary{shapes.geometric, arrows, positioning}

\tikzstyle{startstop} = [rectangle, rounded corners, minimum width=2cm, minimum height=0.75cm, text centered, draw=black, fill=gray!30]
\tikzstyle{process} = [rectangle, minimum width=2cm, minimum height=0.75cm, text centered, draw=black, fill=blue!20]
\tikzstyle{decision} = [diamond, minimum width=2cm, minimum height=0.75cm, text centered, draw=black, fill=yellow!30]
\tikzstyle{arrow} = [thick,->,>=stealth]
\tikzstyle{line} = [draw, -latex']

\title{The Unreasonable Ineffectiveness of Nucleus \\ Sampling on Mitigating Text Memorization}

\author{
  \textbf{Luka Borec\textsuperscript{1}},
  \textbf{Philipp Sadler\textsuperscript{1}},
  \textbf{David Schlangen\textsuperscript{1,2}}
\\
\\
  \textsuperscript{1}CoLabPotsdam / Computational Linguistics\\
    Department of Linguistics, University of Potsdam, Germany\\
  \textsuperscript{2}German Research Center for Artificial Intelligence (DFKI), Berlin, Germany
\\
  \small{
    \textbf{Correspondence:} \href{mailto:firstname.lastname@uni-potsdam.de}{firstname.lastname@uni-potsdam.de}
  }
}

\newcommand{\topp}{\texttt{top\_p} }
\newcommand{\neosmall}{$125$M }
\newcommand{\neolarge}{$350$M }

\begin{document}
\maketitle

\begin{abstract}
This work analyses the text memorization behavior of large language models (LLMs) when subjected to nucleus sampling. Stochastic decoding methods like nucleus sampling are typically applied to overcome issues such as monotonous and repetitive text generation, which are often observed with maximization-based decoding techniques. We hypothesize that nucleus sampling might also reduce the occurrence of memorization patterns, because it could lead to the selection of tokens outside the memorized sequence.
To test this hypothesis we create a diagnostic dataset with a known distribution of duplicates that gives us some control over the likelihood of memorization of certain parts of the training data.
Our analysis of two GPT-Neo models fine-tuned on this dataset interestingly shows that (i) an increase of the nucleus size reduces memorization only modestly, and (ii) even when models do not engage in ``hard'' memorization -- a verbatim reproduction of training samples -- they may still display ``soft'' memorization whereby they generate outputs that echo the training data but without a complete one-by-one resemblance.
\end{abstract}

\begin{figure}[t]
    \centering
    \includegraphics[width=0.49\textwidth]{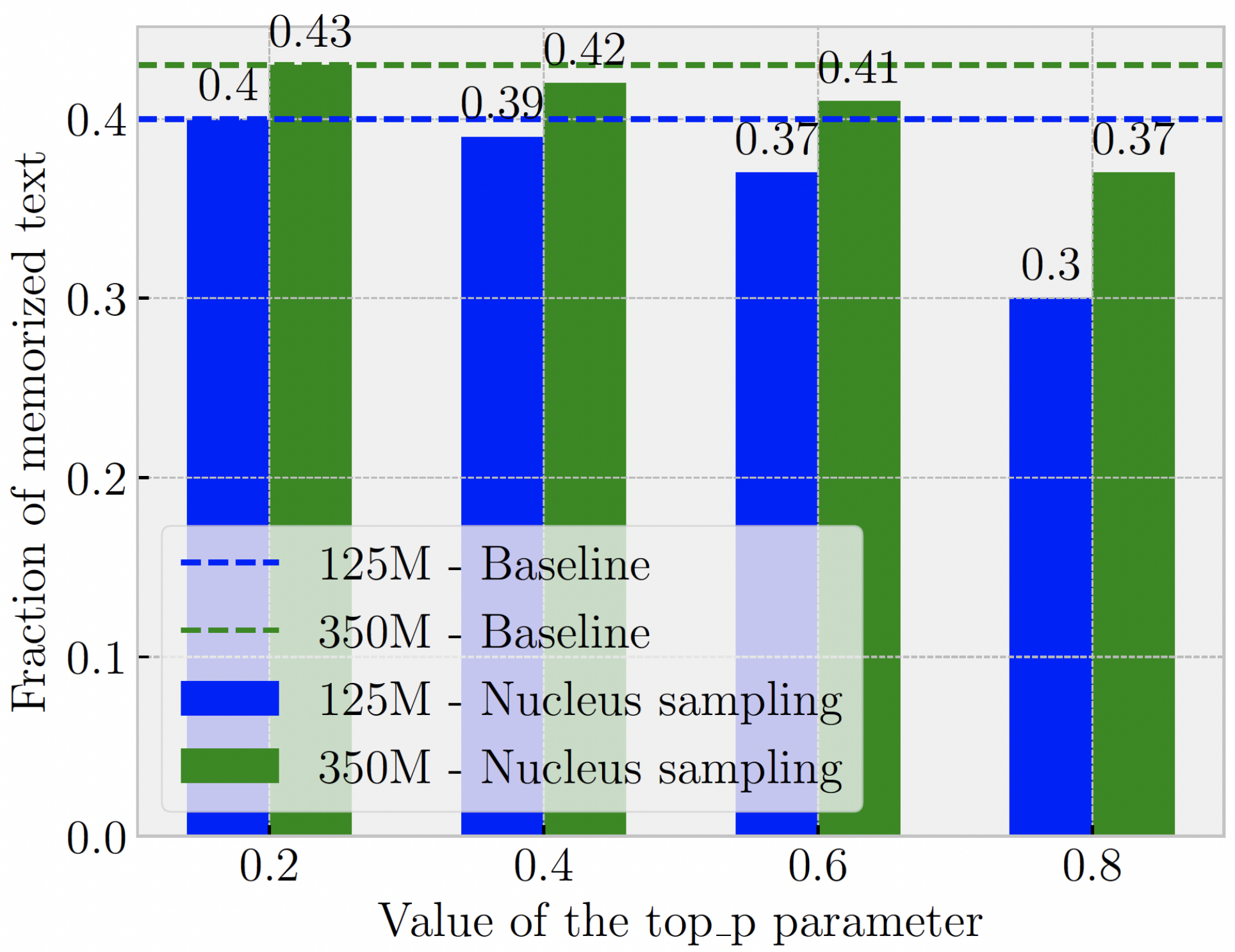}
    \caption{The effect of different \topp values (x-axis) on the fraction of the duplicated texts memorized by the models (y-axis). The \topp parameter determines the maximally considered accumulated probability mass for the output token selection during nucleus sampling. Higher \topp values generally lead to reduced memorization, yet the decrease is less significant than expected. This effect is observed across two models of different model sizes, with the larger model showing a somewhat less pronounced reduction in memorization compared to the smaller model. The dashed lines show the baseline behavior using greedy decoding.}

    \label{fig:nucleus_effect}
\end{figure}

\lstset{
    basicstyle=\ttfamily,
    breaklines=true,
    columns=fullflexible
}

\section{Introduction}

Recent developments in LLMs have led to impressive capabilities in generating human-like text. However, there is growing concern about these models' potential to memorize and regurgitate text from their training data, raising privacy, security, and copyright issues \citep{DBLP:conf/emnlp/0009SC22, DBLP:conf/www/LeeLC023, DBLP:conf/emnlp/KaramolegkouLZS23}. These concerns culminated in a legal dispute between the New York Times and OpenAI which is largely based on the finding that the LLM ``\textit{can generate output that recites Times content verbatim, closely summarizes it, and mimics its expressive style}''\footnote{\url{https://nytco-assets.nytimes.com/2023/12/NYT_Complaint_Dec2023.pdf}, \textit{visited at: 29.05.2024}}.
And indeed \citet{extracting-data} have observed qualitatively that GPT-2 can memorize data from which it was trained, such as HTML pages and logs, and later demonstrated that duplicated texts significantly contribute to memorization when deterministic decoding is at work \citep{quantifying-memorization}. Could the use of a probabilistic decoding technique like nucleus sampling have prevented the lawsuit?

In this paper, we analyze the impact of nucleus sampling \citep{degeneration} on the degree of text memorization. Nucleus sampling is notable for its ability to effectively blend randomness with a focus on likely outcomes. This decoding method operates by sampling from a truncated output distribution (the ``nucleus'') which includes only the highest-probability tokens whose cumulative probability reaches a predefined threshold specified by \topp. 
While the method still focuses on the more probable tokens, it introduces randomness by allowing sampling among the tokens that are otherwise less likely to be generated. 
This makes nucleus sampling a good choice for our study as it aligns with our objectives to explore if and how stochasticity in decoding can mitigate text memorization.

We experiment with a range of nucleus sizes to measure their effects on a model's text memorization behavior (see Figure~\ref{fig:nucleus_effect}). However, quantifying this impact precisely for current very large models is challenging because enumerating duplicates in their training datasets (if they are even accessible) is 
computationally infeasible. To address this, we select a manageable portion of the OpenWebText dataset \citep{Gokaslan2019OpenWeb} and introduce duplicates in a controlled way. This allows us to precisely measure the influence of duplication on memorization, and the degree to which the choice of the decoding strategy can reduce it.

Our findings confirm the previously measured strong correlation between data duplication and memorization  \citep{quantifying-memorization} and deliver new insights about the effects of nucleus sampling: Small nucleus sizes produce effects similar to greedy decoding, and interestingly, even larger nuclei show an ``unreasonable ineffectiveness'' on the mitigation of text memorization, because in cases of peaked distributions a model's memorized token dominates the output distribution, so that even larger nuclei are highly susceptible to generate them. 
Our contributions are as follows:

\begin{enumerate}
    \itemsep-.0em
    \item We create OpenMemText, a diagnostic dataset based on OpenWebText \citep{Gokaslan2019OpenWeb} that contains a controlled number of copies to induce, measure and analyse the memorization behavior of LLMs.
    \item We replicate the results from \citet{carlini2022membership} with two GPT-Neo models \citep{gpt-neo} of different sizes and our results show similar memorization trends with respect to (a) the models' size, (b) the number of duplicates, and (c) the length of the prefix.
    \item We present a comprehensive analysis of the text memorization behavior of the models when using nucleus sampling instead of greedy decoding and find it to be surprisingly ineffective in mitigating text memorization.
\end{enumerate}

    

\section{Related work}
\vspace{-0.2cm}
\paragraph{Text Memorization in Large Language Models.}

\citet{DBLP:conf/fat/BenderGMS21} raised concerns about the magnitude of LLMs, highlighting environmental and accessibility issues, but also noting that these models, much like parrots, tend to repeat the data they have seen during training, leading to issues such as amplifying biases.
%
\citet{DBLP:conf/acl/Magar022} evaluated pre-trained BERT models concerning data contamination and argued that a model's test performance may be inflated by the model's ability to memorize training examples and reproduce them almost verbatim at test time. 
And indeed \citet{DBLP:conf/nips/TirumalaMZA22} found that larger models can memorize large portions of the text without showing overfitting signals. 
\citet{DBLP:journals/corr/abs-2205-10487} argue that the number of data duplicates induces a shift from generalization to memorization.
\citet{DBLP:conf/eacl/HavivCGSGG23} suggest probing for memorized text with specifically constructed English idioms and compare the models' behavior for memorized and non-memorized inputs. \citet{DBLP:conf/nips/ZhangILJTC23} propose counter-factual memorization and measure how the prediction of an LLM changes when specific pieces of information are not shown during training. \citet{DBLP:conf/icml/KandpalDRWR23} confirm that LLMs are sensitive to the number of duplicates seen during training for fact-based question answering and found that deduplication mitigates privacy risks in language models \citep{DBLP:conf/icml/KandpalWR22}. \citet{marone2023dataportraits} introduce Data Portraits, which enable querying of training datasets for membership inference, deduplication, and overlap analysis.


\vspace{-0.2cm}
\paragraph{Decoding Methods for Text Generation.}

Decoding methods transform the probabilistic outputs of language models into readable text. Traditional approaches like greedy decoding follow deterministic rules by choosing the highest probability word at each decision point. Although efficient, text generated in this way is often monotonous and predictable \citep{kulikov2019importance}.  
Sampling-based methods and various decoding heuristics can enhance the diversity and richness of the generated text. \citet{klein-etal-2017-opennmt} propose n-gram blocking to further refine the output quality by preventing the repetitive generation of the same sequence. \citet{DBLP:conf/inlg/GarneauL23} propose an extension to beam search to mitigate hallucinations and omissions. A common decoding technique used with LLMs is temperature sampling \citep{ficler-goldberg-2017-controlling} which adds control over the uniformity of the output distribution, so that a higher temperature leads to likely more versatile outputs because the overall distribution becomes more uniform. 



\section{Memorization Effects in GPT-Neo Models for Greedy Decoding}

\citet{quantifying-memorization} uncovered log-linear relationships between memorization and model size, number of duplicates, and input length, respectively. In particular, they measured the effects of greedy decoding on the memorization behavior of GPT-Neo models using The Pile \citep{gao2020pile} dataset. But they
could only approximate the impact of duplicates due to dataset's unknown duplicate count. Thus, while their study represents one of the most comprehensive quantitative analyses of memorization to date, their findings are based on estimates from their sampled data. In this section, we present the replication of their results using a diagnostic dataset that allows us to measure the amount of text memorization for greedy decoding more precisely.

\subsection{OpenMemText: A Diagnostic Dataset for Text Memorization Research}

\citet{DBLP:conf/nips/BidermanPSSAPR23} has shown that a highly controlled setup is fruitful for the analysis of LLMs and leads to novel insights. Following this paradigm, we create a modified version of the OpenWebText \citep{Gokaslan2019OpenWeb} dataset, an open-source replica of OpenAI's WebText that was used for GPT-2 training. 
OpenWebText contains texts from diverse platforms such as Reddit and news websites. It is $38$ GB uncompressed and consists of over $8$ million curated and \textit{deduplicated} plaintext files each of which represents a separate data point (see Appendix~\ref{appendix:example} for an example data point).
Large datasets present significant challenges in measuring duplicates due to their vast size. However, the deduplicated nature of OpenWebText allows us to manually introduce a known number of duplicates with precise control over their distribution. This enables us to quantify the effect of duplicates in the data on a model's memorization behavior accurately without the computational burden of enumerating duplicates.




\begin{table}[b]
\centering
\begin{tabular}{lll}
\hline
\textbf{Duplicity}  & \textbf{\# Data Points} & \textbf{\# Files} \\ \hline
Zero   &  $\approx500,000$ & $1$ \\
$n-1$ & $280$           & $n \in [2, 30]$            \\
\end{tabular}

\caption{For our analysis, we create a dataset where about $500$K files occur only once and $8120$ samples are duplicated multiple times. As a result, in the majority of cases a data points occurs only once
and we get a balanced distribution concerning the number of copies seen more than once ($2$ times up to $30$ times).}
\label{table:tasks}
\vspace{-.5cm}
\end{table}

To create the dataset in a controlled way, we first sample $0.5\%$ of the OpenWebText files at uniform random which amount to roughly $500$K files. Then we introduce a balanced distribution of duplicates as follows: We select from the files $280$ and duplicate each of them once, so that they appear twice in the dataset. Then we repeat this process by selecting from the remaining files another set of $280$ data points and duplicate them twice, so that they appear three times in the dataset. We repeat this process, each time increasing the duplicate count, until we have files that appear 30 times. This results in approximately $680$K data samples ($4.4$GB) for training, including $180$K duplicates and $500$K files that are not duplicated. We perform the same procedure for the validation set ($1.4$GB) by sampling $0.1\%$ of the OpenWebText files after exclusion of the training samples which resulted in about 400,000 file.

\subsection{Experimental Setup}

First, we ensure that our experimental setup is correct by replicating the results from \citet{quantifying-memorization} with our newly proposed diagnostic dataset.

\paragraph{Model Selection.} For reasons of comparison with the work of \citet{quantifying-memorization} we choose similarly two commonly available GPT-Neo \citep{gpt-neo} models. These models have the same architecture as the GPT-3 \citep{gpt-3} models and were also pre-trained on The Pile \citep{gao2020pile} dataset for over $400$K steps seeing about $420$ billion tokens. For our experimental purposes, we select the \neosmall and \neolarge parameter variants of GPT-Neo model family. Alongside these models, we use the pre-trained GPT-2 as a baseline for the effects of greedy search on the text memorization.

\begin{figure}[t]
    \centering
    \includegraphics[width=0.45\textwidth]{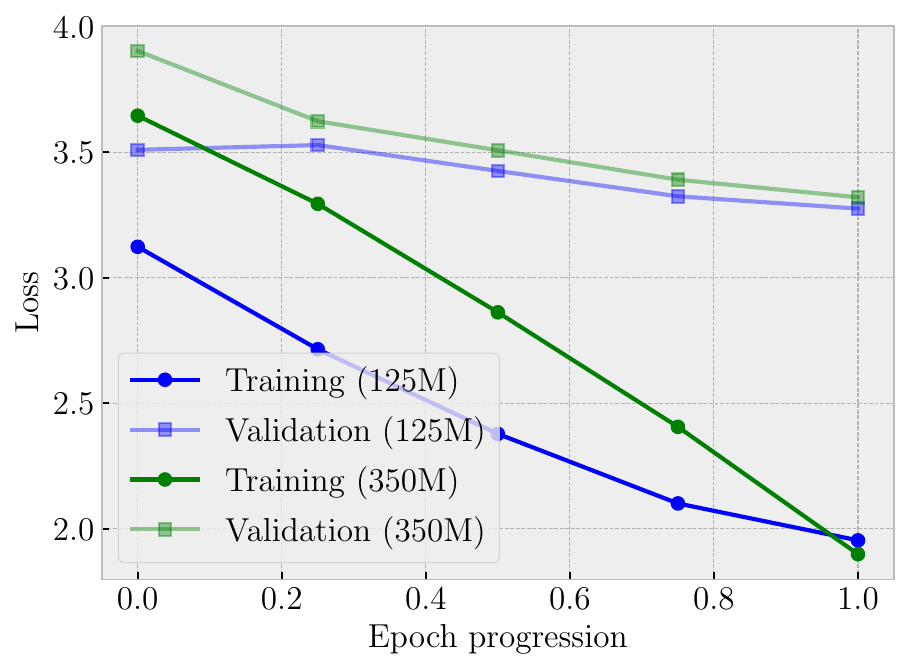}
    \caption{During fine-tuning we measure a consistent decrease in both training and validation loss which indicates that the GPT-Neo models are fitting better to the memorization dataset data over time. }
    \label{fig:loss}
\end{figure}

\paragraph{Model Fine-tuning.} We shuffle the data points in our diagnostic dataset and fine-tune the GPT-Neo models for a single epoch on them. For the \neosmall model we use a batch size of $16$ (distributed across four GPUs), and for the \neolarge model we use a batch size of $4$. We use adaptive learning rate starting at $5e-4$ and employ half floating point precision (fp16) to enhance the fine-tuning efficiency. Based on findings by \citet{mireshghallah-etal-2022-empirical} we specifically target the model's attention heads for fine-tuning and keep the rest of the parameters frozen. The attention heads were found to be the most susceptible to memorization. We argue that a more effective fine-tuning method allows us to better measure how text memorization manifests in the language models compared to less susceptible methods.
%
%
Figure \ref{fig:loss} shows that the fine-tuning method is effective. 


\paragraph{Model Evaluation.} \citet{quantifying-memorization} define memorization as the behavior of a model $f$ to reproduce an exact target string $s$ from the training data $\text{TD}$ when prompted with a certain number of context tokens $p$ (the prefix) of length $\text{len}(s) - k$ such that $f(p) = s$. This behavior can be formalized as:

\begin{equation}
\begin{aligned}
\exists p \colon & \text{len}(p) = \text{len}(s) - k \, \text{and} \\
                 & [p \,||\, s] \in \text{TD} \, \text{and} \\
                 & f(p) = s
\end{aligned}
\label{eq:mem}
\end{equation}

where 

\begin{itemize}
    \itemsep-.0cm
    \item $s$ represents the target string,
    \item $p$ represents the context string with a length of $\text{len}(s) - k$,
    \item $f$ is the model,
    \item $\text{TD}$ denotes the training data for the model $f$,
    \item $[p \,||\, s]$ is the concatenation of the context string $p$ with the target string $s$,
    \item and $f(p) = s$ signifies that the model $f$, when prompted with $p$, produces the string $s$.
\end{itemize}

We use this definition of memorization in our work as well. For instance, if a model’s training dataset contains the sequence ``\textit{Twinkle, twinkle, little star, how I wonder what you are,}'' and given the prefix ``\textit{Twinkle, twinkle, little star,}'' the model outputs ``\textit{how I wonder what you are,}'' this sentence would be considered memorized. 


\begin{figure*}[t]
    \centering
    \includegraphics[width=0.95\textwidth]{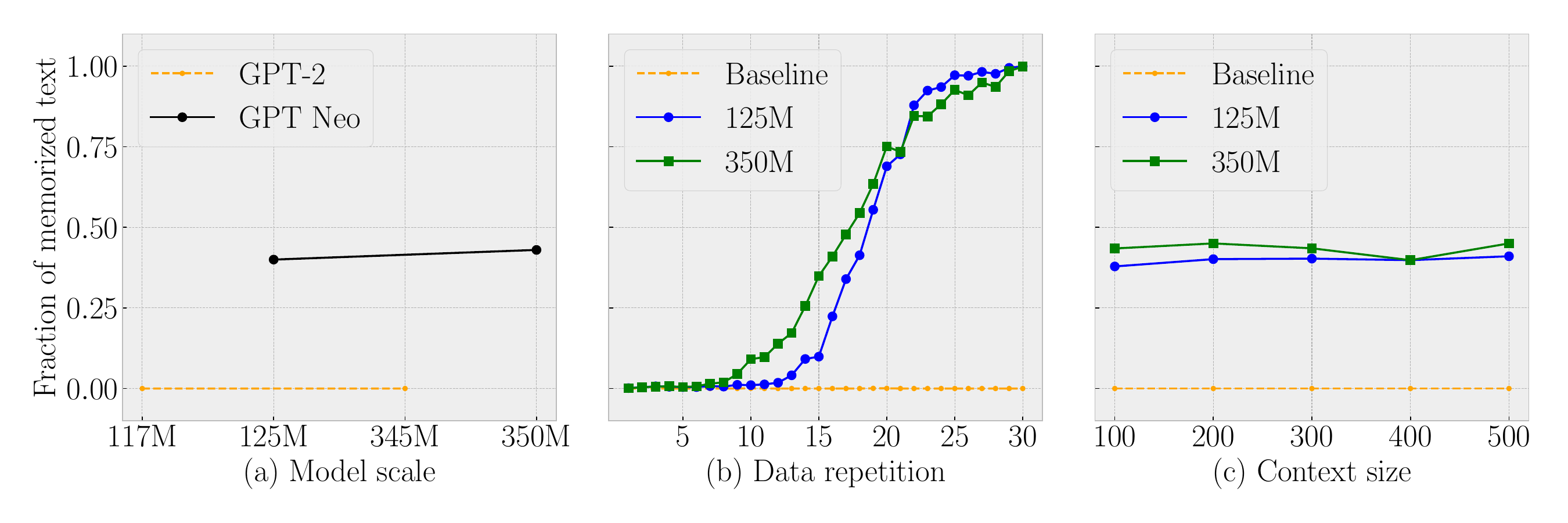}
    \caption{Results from our replication of \citet{quantifying-memorization}. The two fine-tuned GPT-Neo models were compared to non-fine-tuned GPT-2 models of similar sizes using the same prompts. \textbf{(a)} The larger model memorized more of the training dataset than the smaller one. \textbf{(b)} Repeated data in the training set is more likely to be extractable. 
    \textbf{(c)} There is a gradual increase in the extraction of memorized text as the length of input context increases.}
    \label{fig:replication_results}
\end{figure*}

\paragraph{Replication Experiments.} 
For the replication experiments we use all data points from the training dataset with a duplicity greater than zero (see Table~\ref{table:tasks}). For each data point we prompt the model with an experiment specific number of context tokens $p$ and use greedy decoding to generate tokens until an end-of-sentence token or a number of $512$ tokens is produced (note that some samples only contain up to $200$ tokens). We compare the resulting string $s$ with the ground-truth in our training data and count the result as an instance of text memorization in accordance to Equation~\ref{eq:mem}. In particular, we measure the memorization outcomes with respect to the following conditions:




\begin{enumerate}[label=(\alph*)]
    \item \textbf{Model Size:} This experiment explores how model size affects memorization. We use two models containing \neosmall and \neolarge parameters, respectively, and run the memorization experiment with a context length of $p=150$. Our results confirm the findings by \citet{quantifying-memorization} that larger models tend to memorize more as GPT-Neo \neolarge memorized $43\%$ of all duplicated data points whereas the \neosmall parameter model memorized only $40\%$.
    \item \textbf{Data Repetition:} This experiment is conducted in the same way as the one before, but measures the amount of memorization with respect to the number of duplicates. Our trends confirm the original findings by \citet{quantifying-memorization} that more duplicates lead to higher counts of memorized text. Furthermore, we find that the \neolarge parameter model memorizes faster, but both models start to saturate at similar levels.
    \item \textbf{Context Length:} This experiment is conducted as before, but we vary the context length $p$ from 100 to 200, 200 to 300, 300 to 400, and 400 to 500, and over 500 tokens. The scores for each bucket are averaged across all duplicated files belonging to that bucket. While our results somewhat confirm the original paper's findings that an increase in memorization follows an increase in context length, there is a dip at the 300-to-400 length bucket. It is possible that this was caused by small sample sizes for each bucket (70 data points).

\end{enumerate}

Since our results as shown in Figure \ref{fig:replication_results} match those of \citet{quantifying-memorization}, we conclude that our experimental setup works and move on to our nucleus sampling experiment.


\section{Analysing Nucleus Sampling-based Text Memorization Behavior}

This section presents our analysis of text memorization behavior for the fine-tuned GPT-Neo models when using nucleus sampling instead of greedy decoding. In particular, we measure the amount of text memorization of the fine-tuned models under a variety of secondary conditions.

\subsection{The Effect of Duplicates on Text Memorization under Nucleus Sampling}

First, we are interested in the effect of the amount of data duplication on text memorization conditioned on various nucleus sampling thresholds. We conduct the experiments as described for the replication experiments, but with nucleus sampling and different \topp parameters ($0.2,0.4,0.6,0.8$) which determines the size of the nucleus from which the output token is sampled. For our analysis we group the measured amount of memorized text along with the according \topp values.

\begin{figure*}[t]
  \centering
  \includegraphics[width=0.95\textwidth]{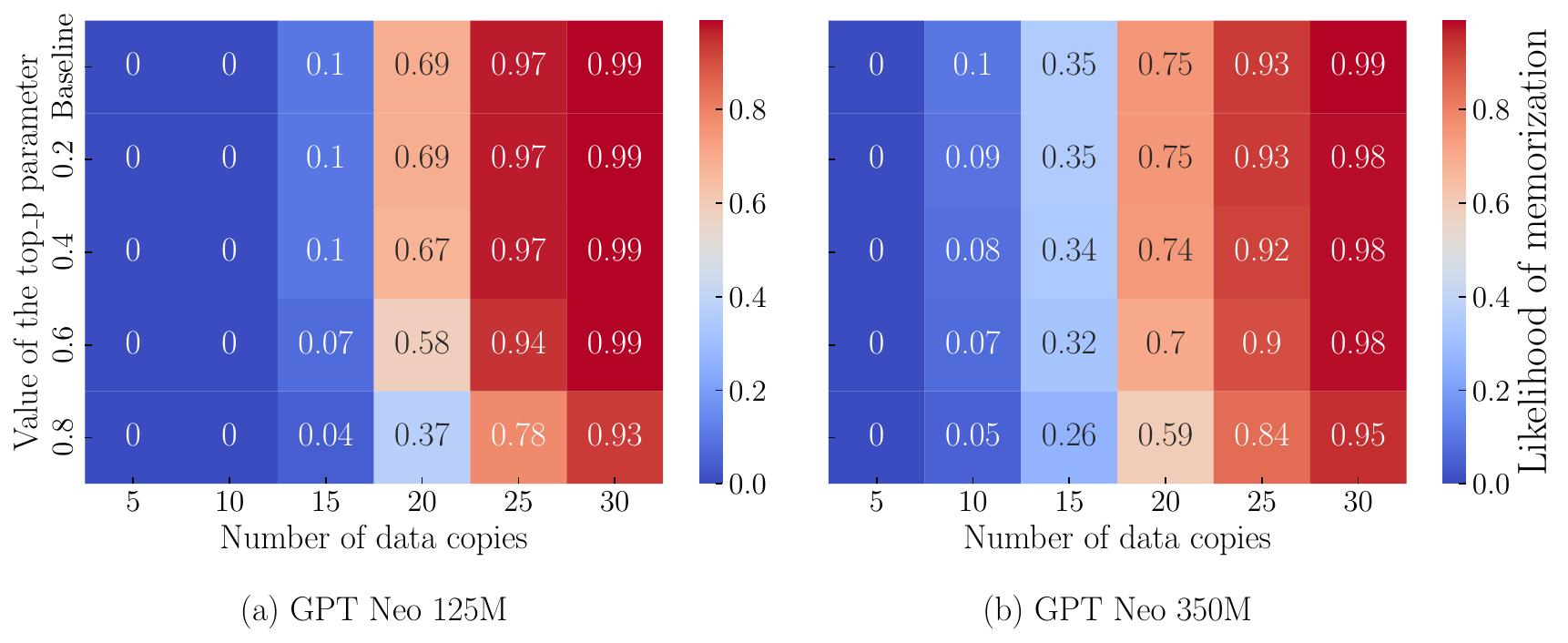}
    \caption{Heatmap illustrating the inverse relationship between \topp parameter values and extracted memorized text, modulated by the number of data repetitions in steps of five. It highlights the unexpected trend that for a high number of data copies, memorization levels remain significant for all \topp values, while fewer data repetitions lead to markedly lower memorization when \topp is increased, reflecting the models' shift from rote memory to learned generalizations.}
  \label{fig:heatmap_a}
\end{figure*}

The resulting heatmap (see Figure~\ref{fig:heatmap_a}) reveals that the larger model consistently shows a higher tendency to memorize across all \topp values. This means that the finding from \citet{quantifying-memorization} that larger models memorize more is also true for nucleus sampling, when all other variables are kept constant. 
Furthermore, we note an intriguing interaction between the duplicate count and the \topp parameter. Especially with high data repetitions ($25$ to $30$ copies) memorization occurs irrespective of the \topp setting. Even with a  $\topp=0.8$ the amount of detected memorized text is nearly equivalent to that of the deterministic greedy search. 

\begin{figure*}[t]
  \centering
  \includegraphics[width=\textwidth]{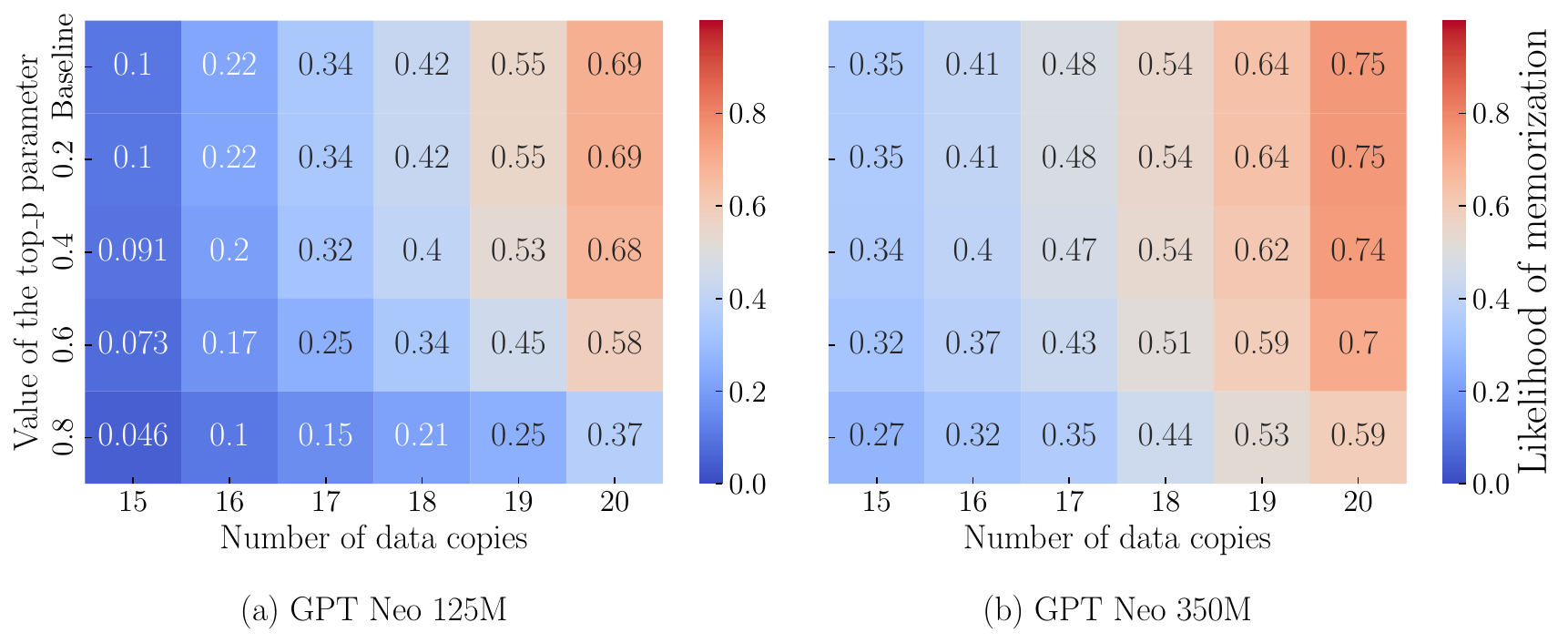}
  \vspace{-0.8cm}
\caption{This more fine-grained view between $15$ to $20$ data copies delineates the ramp-up point where memorization begins to climb sharply and approaches the saturation point where further data addition has diminished effects on memorization rates. This illustrates how, despite increasing \topp values which typically reduce memorization, the presence of high repetition still results in substantial memorization, particularly in the GPT-Neo \neolarge model.}
  \label{fig:heatmap_b}
  \vspace{-.3cm}
\end{figure*}

In contrast, with fewer data copies (up to $20$), increasing the \topp value markedly reduces the amount of memorized content, creating a distinct gap compared to the greedy search which often extracts nearly double the amount.

We conclude that more repetitions allow the models to better internalize sequences, boosting recall. Thus, even with large nuclei, output closely mirrors the training data, making the difference between greedy search and nucleus sampling minimal. However, with fewer data copies, models exhibit reduced memorization, leading to a greater disparity in content retrieval between greedy search and nucleus sampling with larger nuclei.

\vspace{0.2cm}
\textbf{Finding 1:} \textit{At high data repetition, significant memorization occurs across all \topp values in nucleus sampling. However, with lower repetition, lower \topp values lead to higher memorization compared to higher \topp values.}

\subsection{The Emergence of Ramp-up and Saturation Points}

In our analysis we identify stages when a model starts to significantly memorize data from its training set and define these as \textit{ramp-up points}. In addition, we identify \textit{saturation point} as such when further data additions do not significantly improve learning, indicating diminishing returns. 

We find these points prominently illustrated in the middle columns of Figure \ref{fig:heatmap_a}. During the decoding experiments with nucleus sampling, the memorization rates of smaller models significantly ``ramp up'' from $10\%$ at $15$ duplicates to nearly $70\%$ at $20$ duplicates, eventually saturating at $93\%$ at $25$ duplicates. In the larger \neolarge GPT-Neo model, noticeable increases in memorization occur as follows: at $10$ duplicates, memorization stands at $10\%$. This rises to $35\%$ at $15$ duplicates, further escalates to $75\%$ at $20$ duplicates, and peaks at $93\%$ by $25$ duplicates.
We have a closer look at these ramp up points and provides a more detailed view for each duplicate count from $15$ to $20$ in Figure \ref{fig:heatmap_b}. Given this we see that in the case of GPT-Neo \neosmall, memorization remains minimal, with only $1.8\%$ of data memorized up to $12$ data copies. And already at $13$ data copies the amount drastically doubles to $4.1\%$, and doubles again to $9\%$ at $14$ copies. GPT-Neo \neolarge shows a similar pattern. This illustrates how even a single increase in the number of duplicates significantly impacts memorization.

We find that especially at these pivotal \textit{ramp-up points}, where a slight increase in duplicates leads to substantial increases in memorization, employing a larger nucleus size proves effective in reducing text memorization. However, once the models seem to reach a \textit{saturation point}, the efficacy of increasing nucleus size to mitigate memorization diminishes significantly. 

\vspace{0.2cm}
\textbf{Finding 2:} \textit{Higher \topp values reduce memorization significantly at ramp-up points but are much less effective near saturation points where additional data yields diminishing returns.}
\vspace{0.2cm}

A closer look into the \topp values in Figure \ref{fig:heatmap_b} and their effect on memorization rates fosters this finding. When looking at the numbers for the smaller \neosmall GPT-Neo model, then the transition from a more deterministic \topp of $0.2$ to a more stochastic \topp of $0.8$ significantly reduces memorization rates. The memorization decreases from $10\%$ at \topp $0.2$ to $4\%$ at \topp $0.8$ when considering $15$ duplicates, and from $69\%$ to $37\%$ when considering $20$ duplicates. 

These levels can be considered ramp-up points where the difference between \topp $0.2$ and $0.8$ is substantial. However, at $25$ duplicates, where the model appears to be reaching its saturation point, the memorization rates are $97\%$ for \topp $0.2$ and $78\%$ for \topp $0.8$ are showing a lesser though still notable reduction. In the larger \neolarge GPT-Neo model, this trend towards saturation is evident: for data points with $25$ duplicates, the measured text memorization is at $93\%$ under \topp $0.2$ compared to $84\%$ at \topp$ 0.8$.

A possible explanation for this effects is the data density which significantly influence the dynamics of model behavior, especially regarding how quickly saturation points are reached. In datasets abundant with unique items, we would expect the models to experience delayed saturation due to the complexity and infrequency of duplicate data points. Conversely, our diagnostic dataset, rich in multiple copies, likely acts as a ``forced attention'' mechanism. This effect is particularly pronounced in the larger \neolarge GPT-Neo model which due to its higher capacity can better ``incorporate'' the duplicated data points and potentially reach the saturation points more swiftly.

\subsection{The Disturbing Effects of Peak Distributions on Nucleus Sampling}

\begin{figure*} 
  \centering
  \includegraphics[width=\textwidth]{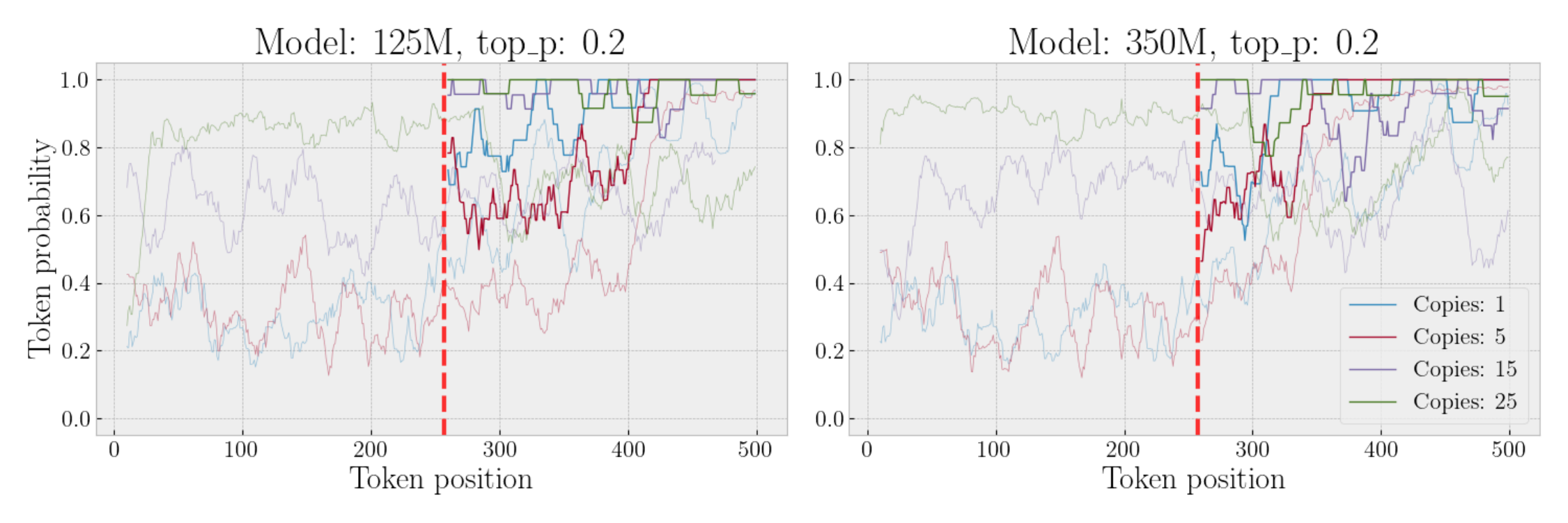}
  \vspace{-1.0cm}
    \caption{The measured token-level probabilities for four randomly sampled data points with an increasing amount of duplicates ($1$, $5$, $15$, and $25$ times) in the training dataset. The thin lines represent the context token probabilities, whereas the bold lines show the probabilities during nucleus sampling with $\topp=0.2$ for an input context length of 250. The horizontal lines on top indicate that a token might be deterministically chosen even for nucleus sampling because its probability exceeds the size of the nucleus.
    }
  \label{fig:top_p_0_2_visualizations}
  \vspace{-0.2cm}
\end{figure*}

We intensify our analysis and have a detailed look on the output distributions of our fine-tuned GPT-Neo models. We select four data points from the diagnostic training set  which appear increasingly often ($1$, $5$, $15$, and $25$ times) and measure the probability of the most likely token to be produced as shown in Figure~\ref{fig:top_p_0_2_visualizations}. 
The results show that the models tend to assign a higher probability to the individual tokens which would lead to an exact continuation of the training text when such texts are seen more often during fine-tuning.

We also examine the differences in token-level probabilities between the tokens used as the context $p$ and those generated by the model. Generated tokens are derived from a subset that the model predicts as most likely for the next position in the sequence. This typically results in higher probabilities for these tokens. In contrast, the probabilities of context tokens can vary widely, as they are not constrained to belong to a sorted group of tokens with cumulative probabilities meet a predefined threshold.
For example, when the nucleus threshold is set to so that $\topp=0.2$, then only tokens (or sometimes even just a single token) whose cumulative probabilities do not exceed the threshold are considered for selection. This effectively excludes other token from being generated. This pattern is illustrated in Figure \ref{fig:top_p_0_2_visualizations}, where such a selection process often occurs for a \topp of 0.2, especially as the number of duplicate tokens increases.

We conclude that using low \topp values is often less effective for mitigating memorization issues. This occurs because snippets that the model has memorized, which usually have high token-level probabilities, tend to dominate the selection process. When these probabilities exceed the \topp threshold, the decoding process essentially becomes deterministic because the nucleus can consist of only a single token. This is problematic especially when the objective is to mitigate memorization constraints.
This can even happen for higher \topp values, such as $0.4$ (see Appendix~\ref{appendix:evaluation}).

\vspace{0.2cm}
\textbf{Finding 3:} \textit{Models that strongly memorize texts assign very high  probabilities to single tokens so that even nucleus sampling becomes deterministic. This happens when the token's probability exceeds the \topp threshold, so that nucleus to sample from contains only a single candidate token.}

\subsection{The Emergence of ``Soft'' Memorization} 

In the previous analysis we mainly considered text memorization as defined under Equation~\ref{eq:mem} (verbatim memorization) i.e. when every generated token for some context can be found in the training dataset following the same output. However, we argue that measuring memorization in terms of \textit{degrees} rather than binaries would be helpful. 

Inspired by \citet{10.1162/tacl_a_00567} who propose to measure the novelty of generated text with n-grams, we suggest to use an n-gram overlap metric (BLEU,  \citet{DBLP:conf/acl/PapineniRWZ02}) as a weaker, but still meaningful constraint to measure memorization. We again sampled continuations given prefixes from the duplicated material and then measured the overlap of the predicted with the actual continuations, using BLEU-4. To ensure that the scores are not inflated, the initial $250$ tokens used to prompt the model are excluded, focusing solely on the completion. 
An interesting observation from the results in Table~\ref{table:bleu} is the positive correlation between the number of duplicated data and the measured BLEU-4 scores, especially a very high BLEU-4 score for samples represented $20$ and $30$ times. This trend suggests a ``soft memorization'' behavior of the models. A possible explanation is that a higher number of data copies leads the models to alternate between recalling memorized and novel tokens, rather than directly reproducing memorized content. This finding echoes on a recent concerns on ``a false sense of privacy'' when verbatim memorization is not recognized \citep{DBLP:conf/inlg/IppolitoTNZJLCC23,DBLP:conf/fat/BrownLMST22}.


\vspace{0.2cm}
\textbf{Finding 4:} \textit{ Data with many duplicates leads to abnormally high BLEU scores, indicating ``soft memorization'' whereby models alternate between recalling memorized and novel tokens, resulting in outputs that closely resemble the training data without being exact copies.}

\section{Conclusion}

\begin{table}[t]
    \centering
    \small 
    \renewcommand{\arraystretch}{1.2} 
    \begin{tabular}{|c|c|c|c|c|c|}
    \hline
    \multirow{2}{*}{Model} & \multirow{2}{*}{\topp} & \multicolumn{4}{c|}{Number of copies} \\
    \cline{3-6}
    & & 1 & 10 & 20 & 30 \\
    \hline
    \multirow{4}{*}{Neo 125M} & 0.2 & 0.02 & 0.24 & 0.40 & 0.84 \\
    & 0.4  & 0.01 & 0.26 & 0.44 & 0.84 \\
    & 0.6  & 0.01 & 0.26 & 0.37 & 0.84 \\
    & 0.8  & 0.00 & 0.27 & 0.34 & 0.71 \\
    \hline
    \multirow{4}{*}{Neo 350M} & 0.2 & 0.01 & 0.28 & 0.42 & 0.74 \\
    & 0.4  & 0.01 & 0.28 & 0.44 & 0.76 \\
    & 0.6  & 0.02 & 0.28 & 0.40 & 0.73 \\
    & 0.8  & 0.02 & 0.27 & 0.40 & 0.67 \\
    \hline
    \end{tabular}
    \caption{BLEU-4 scores for non-verbatim memorized outputs, considering both the \topp value and the duplicate count of the texts within the training dataset.
    }
    \label{table:bleu}
    \vspace{-0.5cm}
\end{table}

We created a diagnostic dataset to measure the memorization behavior of two Neo-GPT models more precisely than previous work \citep{quantifying-memorization} that relied on an estimate of duplicates in the training data. Given this we fine-tuned the GPT-Neo models on our dataset and confirmed with our replication experiments the other results under greedy decoding. With this experimental setup we analysed the language models productions when nucleus sampling is used for decoding. 

The results show that for models with strongly memorized texts low \topp values in nucleus sampling converge to greedy decoding. We note that even the experiments using large \topp values often fail to substantially mitigate memorization. This at the first glance ``unreasonable ineffectiveness'' of nucleus sampling to mitigate text memorization is mostly caused by high peak distributions -- specifically, when a single token's probability exceeds the cumulative threshold set by the nucleus size, causing nucleus sampling to operate deterministically. Larger nucleus sizes only modestly mitigate memorization, and even when outputs are not exact reproductions, we find that n-gram overlap scores indicate a ``soft memorization'' phenomena.

In further work we will explore the impact of other duplicate distributions in the training dataset on the memorization behavior. Furthermore, more research is needed to confirm if the strategy of fine-tuning the attention heads will generalize to less susceptible methods like adapter-based or full-model fine-tuning and to even bigger models.


\section{Limitations}

\paragraph{Limitations on the range of chosen \topp values.} Our analysis evaluated a spectrum of \topp values: $\{0.2, 0.4, 0.6, 0.8 \}$. Although this chosen range is sufficient to make the presented observations, it is not exhaustive. Text generation tasks that demand high precision and do not necessarily value lexical diversity, such as code generation, allow for relatively low \topp values to be efficient. This is evident in the case of \citet{alphacode}, who, in their experiments with a code generation system that solves competitive programming problems, used \topp values starting from $0.5$ and did not see significant changes in performance beyond $0.8$. Nevertheless, an interesting addition to our experiments would be \topp values of $0.9$ and $0.95$, as proposed by \citet{degeneration}, who demonstrated that these values increase the lexical diversity of generated texts as measured by Self-BLEU \citep{self-bleu}, a metric that evaluates diversity by comparing generated text samples from the same model.



\paragraph{Limitations on model sizes.} Our study covered language models of size and capability that show comparable behaviors to those chosen by \citet{quantifying-memorization}. Nevertheless, we were limited by resource constraints and featured primarily smaller models. An interesting addition would be to use low-rank adapters (LoRA) \citep{hu2021loralowrankadaptationlarge} to apply our presented analysis to large-scale models with billions of parameters as they become publicly available in the future.


\FloatBarrier

\paragraph*{Supplementary Materials} The source code is available at \url{https://github.com/lukaborec/memorization-nucleus-sampling}. We published the OpenMemText dataset at \url{https://doi.org/10.5281/zenodo.13318542}.

\section*{Acknowledgements}

This work was partially funded by the Deutsche Forschungsgemeinschaft (DFG, German Research Foundation) – 423217434 (“RECOLAGE”) grant. 

\bibliography{custom}
\bibliographystyle{acl_natbib}

\clearpage

\appendix

\section{Appendix}

\subsection{Hardware Specifications} 
\label{appendix:hardware}

The experiments were performed on a system equipped with four NVIDIA GeForce GTX 1080 Ti GPUs, 250 GB of RAM, and 12 Intel(R) Xeon(R) CPU E5-2650 v4 @ 2.20GHz cores.

\subsection{Dataset Creation Details}
\label{appendix:dataset}

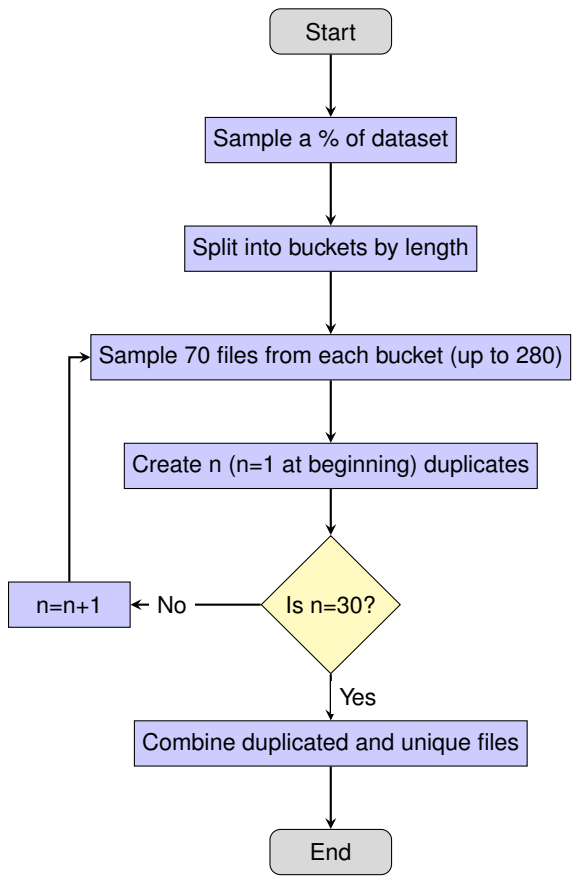
\begin{figure}[h]
    \centering
    \begin{tikzpicture}[node distance=1.8cm, every node/.style={fill=white, font=\sffamily, scale=0.8}, align=center]

    \node (start) [startstop] {Start};
    \node (sample) [process, below of=start] {Sample a \% of dataset};
    \node (bucket) [process, below of=sample] {Split into buckets by length};
    \node (sample70) [process, below of=bucket] {Sample 70 files from each bucket (up to 280)};
    \node (duplicate1) [process, below of=sample70] {Create n (n=1 at beginning) duplicates};
    \node (loop) [decision, below of=duplicate1, yshift=-0.5cm] {Is n=30?};
    \node (increment) [process, left of=loop, xshift=-2.5cm] {n=n+1};
    \node (finish) [process, below of=loop, yshift=-0.5cm] {Combine duplicated and unique files};
    \node (end) [startstop, below of=finish] {End};

    \draw [arrow] (start) -- (sample);
    \draw [arrow] (sample) -- (bucket);
    \draw [arrow] (bucket) -- (sample70);
    \draw [arrow] (sample70) -- (duplicate1);
    \draw [arrow] (duplicate1) -- (loop);
    \draw [arrow] (loop) -- node[anchor=east] {No} (increment);
    \draw [arrow] (increment) |- (sample70);
    \draw [arrow] (loop) -- node[anchor=west] {Yes} (finish);
    \draw [arrow] (finish) -- (end);

    \end{tikzpicture}
    \caption{The dataset creation process depicted as a flowchart. We first sample a percentage of the overall data. Then we split them into buckets by different lengths. From each bucket we sample 70 files repeatedly until we have chosen 280 files. For these chosen file we create duplicates respectively.}
    \label{fig:flowchart}
\end{figure}

To ensure uniformity across different file lengths and facilitate the successful execution of our experiment on input context length, during the initial sampling of the dataset we made sure that the dataset consisted of equal parts texts of lengths up to 200 tokens, 200 to 300 tokens, 300 to 400 tokens, and over 400 tokens. We then sampled 70 files from each bucket, combining them to form the 280 files used for duplication. Figure \ref{fig:flowchart} shows the step-by-step process.

\vspace{0.5cm}

\subsection{Example Data Point}
\label{appendix:example}

An example of a randomly chosen data point showing the tone and the style of the dataset. The text is shown as it appears in the text file, i.e., full length, with the punctuation intact.

\vspace{0.5cm}

\begin{tcolorbox}[colback=white, colframe=black, width=0.45\textwidth]
\begin{small}
\begin{lstlisting}
Came home today to find a package in my mailbox (giggidy). Opened it up to find two nicely wrapped presents. The first one I opened felt like a movie (I love movies) so I eagerly tore off the packaging to find Amelie. A movie I've heard about but have yet to watch. Attached was a note saying it was my Santa's favorite movie and I should watch it, too. I plan on it, Santa, I plan on it.

Then I saw the more oddly shaped package and sat in confusion for a while. I decided to open it right away instead of waiting for Christmas. Upon ripping the wrapping paper off, I saw a Doctor Who TARDIS monitor mate. I'm super excited to use it at work. I haven't decorated my new office yet and this will be perfect!

Thank you, Santa!
\end{lstlisting}
\end{small}
\end{tcolorbox}

\subsection{Training Details}
\label{appendix:training}

We assess the fine-tuning effectiveness of the GPT-Neo models by monitoring loss and perplexity. We notice a consistent decrease in both training and validation loss which indicates that the models are fitting better to the training data over time. However, the validation loss decreases significantly slower than the training loss. This is expected given the abundance of duplicates in the training dataset which the models are overfitting to. As with the loss, Table \ref{tab:perplexity} shows a discrepancy between the training and validation perplexities, reinforcing the earlier assumption of the models overfitting to the duplicates.

\begin{table}[h]
\centering
\begin{tabular}{|c|c|c|}
\hline
\textbf{Model} & \textbf{Training} & \textbf{Validation} \\
\hline
GPT Neo \neosmall & $26.44$ & $7.05$ \\
\hline
GPT Neo \neolarge & $27.66$ & $6.67$ \\
\hline
\end{tabular}
\caption{Calculated perplexities of the fine-tuned models for training and validation splits.}
\label{tab:perplexity}
\end{table}

\twocolumn[\subsection{Evaluation Details}\label{appendix:evaluation} The following figure shows the variation of word-level probabilities in four randomly sampled texts appearing $1$, $5$, $15$, and $25$ times in the training dataset. In nucleus sampling, if the probability of a single token exceeds the size of the nucleus (parameterized by \topp), the entire probability distribution is assigned to that single token while all other tokens are discarded. This seems to happen often at low \topp values and especially so for sentences with a large number of repetitions.\vspace{.2cm}]

\includegraphics[width=.95\textwidth]{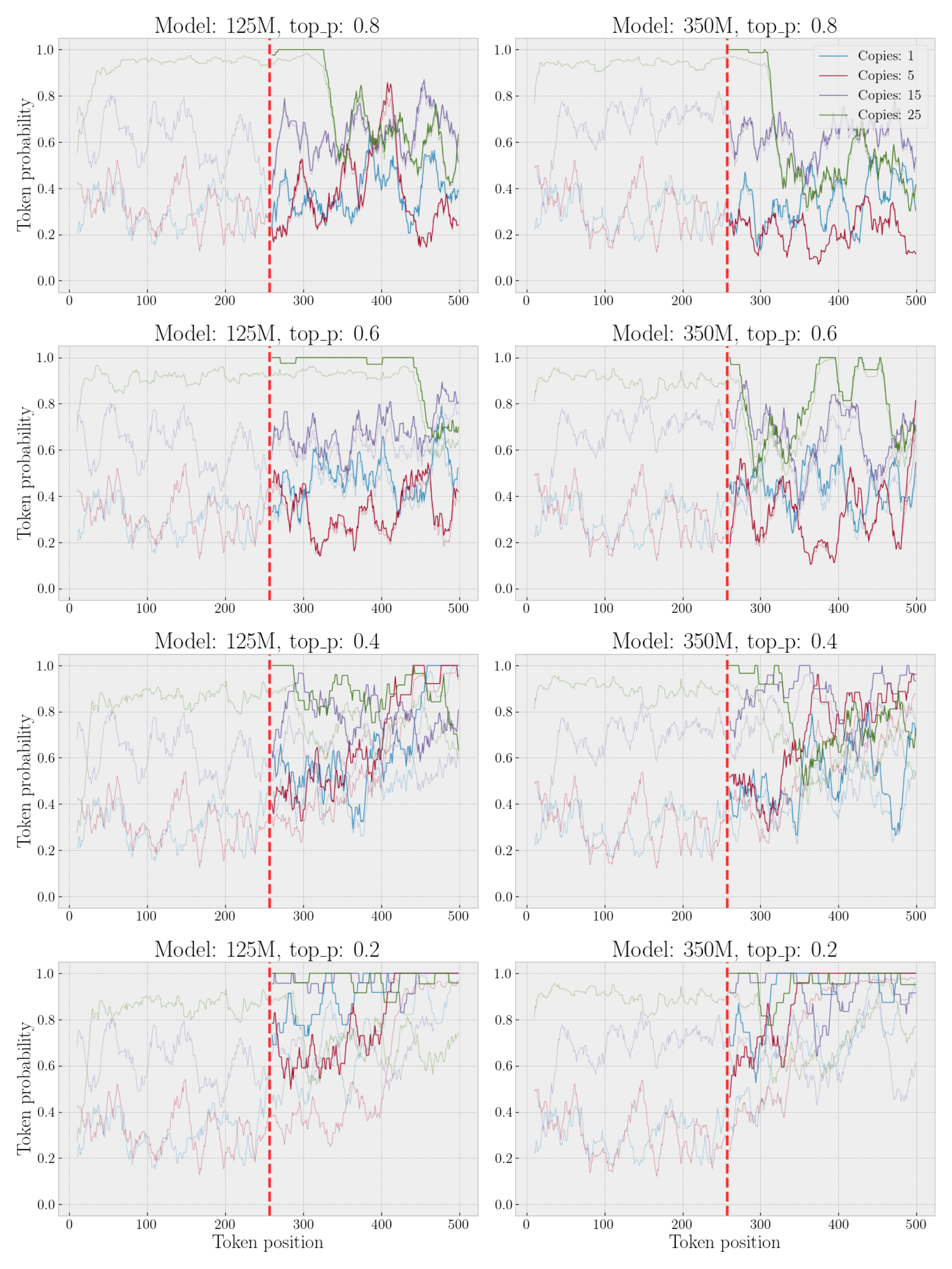}

\end{document}